\documentclass[letterpaper]{article} 
\usepackage{aaai2026}  
\usepackage{times}  
\usepackage{helvet}  
\usepackage{courier}  
\usepackage[hyphens]{url}  
\usepackage{graphicx} 
\usepackage{natbib}  
\usepackage{caption} 
\usepackage{algorithm}
\usepackage{algorithmic}

\usepackage{newfloat}
\usepackage{listings}
\DeclareCaptionStyle{ruled}{labelfont=normalfont,labelsep=colon,strut=off} 
\floatstyle{ruled}
\newfloat{listing}{tb}{lst}{}
\floatname{listing}{Listing}
\nocopyright

\usepackage{amsmath}
\usepackage{amssymb}
\usepackage{booktabs}
\usepackage{xcolor}
\usepackage{tikz}
\definecolor{gold}{HTML}{BD820B}
\definecolor{silver}{HTML}{909090}
\definecolor{bronze}{HTML}{9A5F26}
\definecolor{pred}{rgb}{1, 0.3, 0.3} 
\usepackage{color, colortbl}
\definecolor{Gray}{gray}{0.95}

\newcommand*\circledd[1]{\tikz[baseline=(char.base)]{
            \node[shape=circle,draw,inner sep=0.15pt] (char) {#1};}}      

\newcommand{\name}{\textsc{GeCo2}} 
            
\newcommand{\first}[1]{%
    {#1\raisebox{0.8pt}{\footnotesize \color{gold} \circledd{1}}}%
}
\newcommand{\second}[1]{%
    {#1\raisebox{0.8pt}{\footnotesize \color{silver} \circledd{2}}}%
}
\newcommand{\third}[1]{%
    {#1\raisebox{0.8pt}{\footnotesize \color{bronze} \circledd{3}}}%
}

\title{Generalized-Scale Object Counting with Gradual Query Aggregation}
\author{
    Jer Pelhan, Alan Lukežič, Matej Kristan
}
\affiliations{
Faculty of Computer and Information Science, University of Ljubljana, Slovenia
\\\tt\small jer.pelhan@fri.uni-lj.si}

\usepackage{bibentry}

\begin{document}

\maketitle

\begin{abstract}
  Few-shot detection-based counters estimate the number of instances in the image specified only by a few test-time exemplars. 
A common approach to localize objects across multiple sizes is to merge backbone features of different resolutions.
Furthermore, to enable small object detection in densely populated regions, the input image is commonly upsampled and tiling is applied to cope with the increased computational and memory requirements. Because of these ad-hoc solutions, existing counters struggle with images containing diverse-sized objects and densely populated regions of small objects.
We propose \name{}, an end-to-end few-shot counting and detection method that explicitly addresses the object scale issues. 
A new dense query representation gradually aggregates exemplar-specific feature information across scales that leads to high-resolution dense queries that enable detection of large as well as small objects.
\name{} surpasses state-of-the-art few-shot counters in counting as well as detection accuracy by $\sim$10\% while running $\sim$3$\times$ faster at smaller GPU memory footprint.
Code: https://github.com/jerpelhan/GECO2.
\end{abstract}

\section{Introduction}
 
Few-shot object counters estimate the number of objects, whose category was not observed in training,
using only a few annotated exemplars.
Driven by benchmarks~\cite{famnet,counting-detr}, 
the research has initially focused on global counters, 
which 
report counts as summation over an image-wide estimated density map~\cite{Liu_2022_BMVC,djukic_loca}. 
The continuous density is well suited for crowded regions, but cannot provide exact object locations, which makes the counts unexplainable and unsuitable for applications like biomedical analysis~\cite{zavrtanik2020segmentation}, where precise object positions and sizes estimates are required.

Detection-based counters, which produce per-object bounding boxes thus emerged. 
The current state-of-the-art is based on DETR-like~\cite{detr} detection paradigms with fixed-set pre-trained object queries~\cite{counting-detr}, region proposals~\cite{dave,pseco} or dense object query maps~\cite{geco}.
\begin{figure}[h!]
    \centering
    \includegraphics[width=\linewidth]{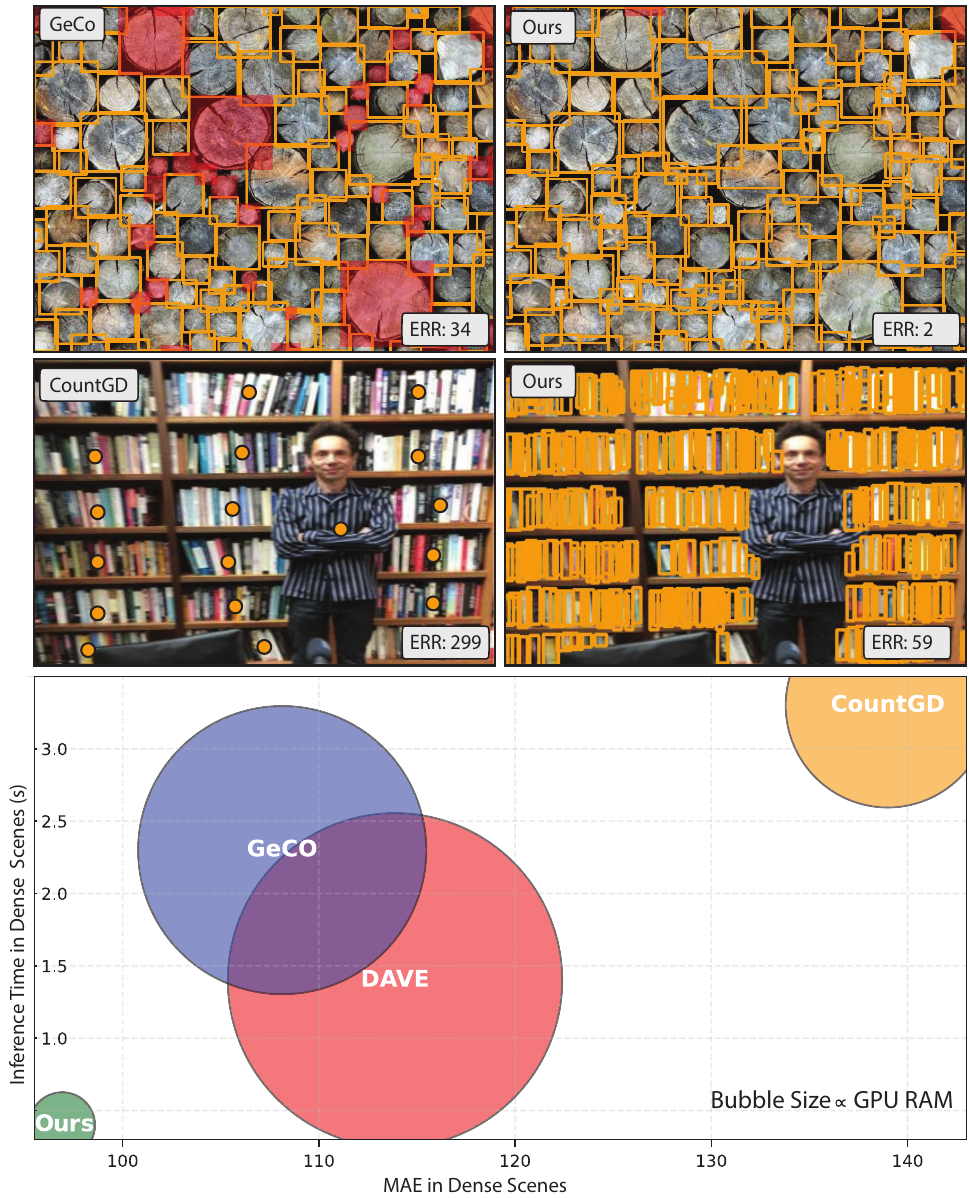}
    \caption{GeCo~\cite{geco} struggles with diverse object scales and CountGD~\cite{countgd} fails in dense scenes with small objects. \name{} overcomes these issues by the new scale-generalized dense queries and achieves superior accuracy, fast inference and a substantially lower memory footprint.}\label{fig:motivation}
\end{figure}
However, these struggle with small objects in dense regions
(Figure~\ref{fig:motivation}).
A major cause is the backbone feature resolution reduction, which causes several objects to occupy the same latent pixel.
A common solution is to upsample the input image~\cite{geco,Liu_2022_BMVC,countgd}, which however is limited by the GPU memory.
To achieve sufficient resolution, tiling is employed~\cite{Liu_2022_BMVC,countgd}, i.e., running inference on sub-images and reassembling the results.
Which leads to several issues: 
(i) exemplars cut by tiles cannot be easily extracted, 
(ii)  
transformer-based backbones yield tile-specific features incompatible with exemplars on tiles with different visual content
and 
(iii) detection accuracy for objects at tile borders is low, often leading to double detections at tile reassembly.  
Moreover, since the scaling-based methods do not consider objects at multiple resolutions, their performance degrades when the object sizes in the image vary substantially (see Figure~\ref{fig:motivation}).

We propose a new \underline{Ge}neralized-Scale \underline{Co}unter \name{}, that explicitly addresses the aforementioned challenges of diverse object sizes and dense regions.
The core innovation is high-resolution dense query map construction, where prototypes are formed at each scale and query interactions occur independently at each scale, followed by cross-scale aggregation. 
While multi-scale features are widely used in vision, the proposed gradual high-resolution query construction through scale-specific encoding with class-defining exemplars on each separate scale is unique and outperforms the classical multi-scale formulations as demonstrated in experiments.
In contrast to the current state-of-the-art, \name{} thus avoids the need for the ad-hoc heuristic input upscaling and tiling for small object detection, 
enjoys a low memory footprint, 
and maintains excellent cross-scale detection capability.

\name{} outperforms all few-shot density- and detection-based counters on the counting benchmark~\cite{famnet} by $\sim$25\% RMSE. 
It further delivers superior detection accuracy, surpassing the best method by 8\% AP. 
On detection-field-adapted FSCD-LVIS~\cite{counting-detr}, \name{}
outperforms state-of-the-art by 15\% MAE in counting and by 23\% AP in detection. 
On the recent multi-class benchmark~\cite{mcac}, it 
dominates by
25\% MAE advantage, demonstrating excellent discrimination capabilities.
Notably, \name{} runs at least 3$\times$ faster than the current detection-based counters, while using less than one-third of their GPU memory, underscoring its practical applicability.


\section{Related work}
 
Traditional object counters originate from object detectors trained for predefined categories such as vehicles~\cite{vehicle-counting}, cells~\cite{cell-counting-detection}, and people~\cite{crowd-counting}, and rely on extensively annotated datasets.
Few-shot counting, introduced with FSC147~\cite{famnet}, shifts the paradigm by adapting to arbitrary and unseen categories at test time.
Few-shot counters primarily follow two paradigms: (i) density map regression and (ii) detection-based counting. 
Early density-based methods, like FamNet~\cite{famnet} employed a Siamese network in combination with test-time fine-tuning,
BMNet+~\cite{Shi_2022_CVPR} formulated similarity metric learning, while SAFECount~\cite{you2023few} introduced a new feature enhancement module.
CounTR~\cite{Liu_2022_BMVC} employed an attention mechanism for exemplar-image feature interaction and introduced test-time normalization to adjust the total counts by the predicted counts within the exemplar bounding boxes.
Furthermore, when the given exemplars are small, indicating the presence of small objects, it divides the input image into sub-images, on which inference is run separately. Both test-time ad-hoc solutions lead to notable performance gains.
LOCA~\cite{djukic_loca} improved generalization and performance by iteratively enhancing object prototypes with shape and appearance with attention.

Detection-based methods, in contrast, generate discrete bounding boxes, offering more informative outputs. 
DETR-based counters~\cite{counting-detr,countgd} rely on a fixed number of object queries (e.g., 600), limiting the counting capacity and forcing usage of ad-hoc tiling for dense scenes, which in practice restricts these counters to global count prediction, as detection merging is unfeasible in dense scenes. 
CountGD~\cite{countgd} introduces the SAM-based test-time normalization, a refined global count normalization based on the number of detections within the SAM~\cite{sam} segmentation mask, which is inferred from exemplar bounding boxes. 
This normalization prevents  reporting object location. 
A recent detection-based counter GeCo~\cite{geco} introduced a dense non-parametric prototype formulation, removing the fixed query constraint. 
However, to cope with small objects, these methods require input upsampling, which significantly increases inference time and memory usage due to the high cost of self-attention over dense representations.

Object scale variation is another fundamental challenge in few-shot counting, as areas of single objects vary from 20 pix$^2$ to 50.000 pix$^2$~\cite{famnet}. Most state-of-the-art approaches~\cite{Liu_2022_BMVC,you2023few,Shi_2022_CVPR} do not address scale and regress on features from a single resolution. 
Other approaches provide only partial solutions. 
Famnet~\cite{famnet} scales the exemplar features, i.e., template, to obtain multiple correlation maps. 
To cope with memory limitations, 
LOCA~\cite{djukic_loca}, DAVE~\cite{dave} and GeCo~\cite{geco} operate on low-resolution features, while GeCo~\cite{geco} upsamples the output using higher-resolution features.
In contrast, our \name{} introduces a new object query map encoder that comfortably scales over different resolutions and proposes a gradual latent query construction that leads to improved detection performance at a lower memory footprint.

\section{\name{}}
\begin{figure*}[h!]
    \centering
    \includegraphics[width=\linewidth]{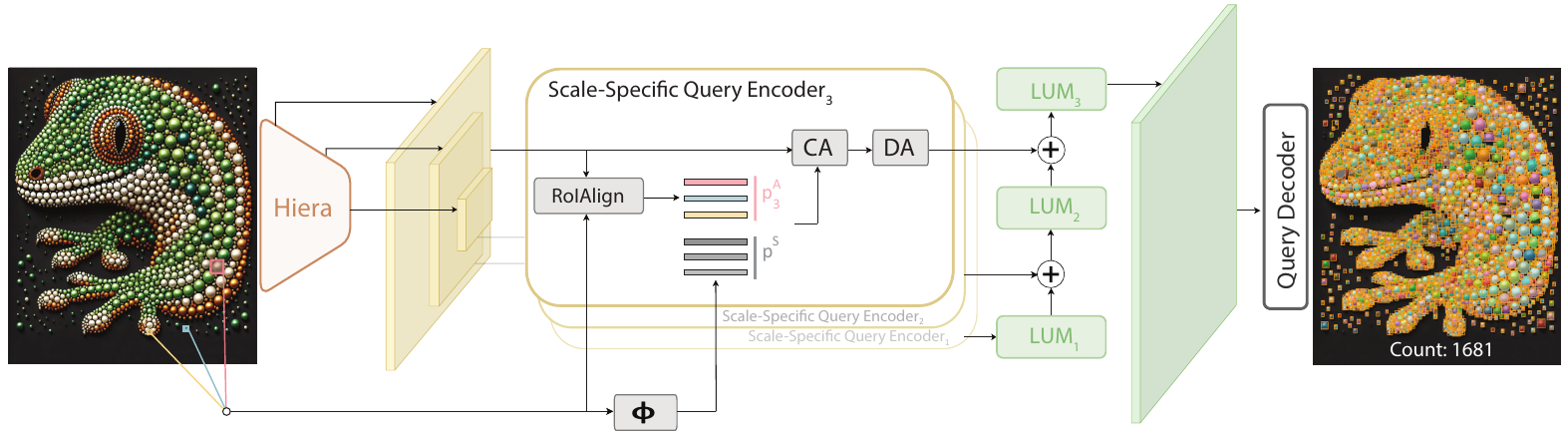}
    \caption{The architecture of the proposed  generalized-scale detection-based object counter \name{}. }
    \label{fig:architecture}
\end{figure*}

Given an image \( I \in \mathbb{R}^{H_0 \times W_0 \times 3} \) 
and a set of \( k \) exemplar boxes \( \boldsymbol{B}^{\text{E}} = \{ \mathbf{b}_i \}_{i=1:k} \), 
the goal is to detect all target instances in \( I \) as a 
set of boxes 
\( \boldsymbol{B}^{\text{P}} = \{ \mathbf{b}_j \}_{j=1:N} \), 
and their count estimated as \( N = |\boldsymbol{B}^{\text{P}}| \).
To cope with dense object regions, we cast the detection problem as dense query map construction~\cite{geco}, which is decoded into the set of final bounding boxes. 
There is a key difference between~\cite{geco}, which predicts a low-resolution query map and upscales it with high-resolution backbone features at the decoding stage, and our approach. 
In particular, our approach directly builds a high-resolution generalized-scale dense query map, exploiting exemplars at multiple scales and avoiding the need for image upscaling.

The proposed \name{} architecture is outlined in Figure~\ref{fig:architecture}. 
Hiera backbone~\cite{hiera} extracts multi-resolution features $\mathbf{f}^{I} \in \{\mathbf{C}_\text{L} \}_{\text{L = 1, 2, 3}}$, where $\mathbf{C}_L \in \{\mathbb{R}^{h_L \times w_L \times d}\}$, $h_L = H_0/r_L$, $w_L = W_0/r_L$, $r \in \{16, 8, 4\}$ and $d$ is number of feature channels.
Each scale is processed separately by a scale-specific query encoder, employing scale-specific appearance and shape prototypes. 
A cross-scale query aggregation module then gradually transforms and merges scale-specific dense queries into a generalized single-scale dense query map, from which object detections are inferred. 
The detected instances are further refined by the SAM2~\cite{sam2} segmentation head. \name{} can thus be seen as a new SAM2 head, adding a counting capability to the SAM2 swiss-knife backbone.

\subsection{Scale-Specific Query Encoder}  \label{sec:prototype_and_query}
 
The task of the scale-specific query encoder is to extract 
exemplar-guided object presence information for the objects corresponding to the scale $L$ of the considered backbone features $\mathbf{C}_\text{L}$. 
To focus on the relevant object types, scale-specific object appearance and object shape prototypes are extracted from the exemplars.
The object shape prototypes $\mathbf{p}^{S} \in \mathbb{R}^{k \times d}$ are constructed from their exemplar width and height 
$(W_{\mathbf{b}_i},H_{\mathbf{b}_i})$
in the original image resolution, i.e., $\mathbf{p}_{i}^{S}=\Phi([W_{\mathbf{b}_i},H_{\mathbf{b}_i}])$, where $\Phi(\cdot)$ is a trainable MLP, as in~\cite{djukic_loca}, with same parameters shared across all scales.

In addition to shape, the appearance prototypes $\mathbf{p}^{A}_{L} \in \mathbb{R}^{k \times d}$ are extracted separately at each scale by the RoiAlign~\cite{roipooling} operation from their bounding boxes. 
The two sets of prototypes ($\mathbf{p}^{A}_{L}$ and $\mathbf{p}^{S}$) are concatenated, forming scale-specific prototypes $\mathbf{p}_{L} \in \mathbb{R}^{2k \times d}$. 
 The scale-specific dense object queries map $\mathbf{Q}_L \in \{\mathbb{R}^{h_L \times w_L \times d}\}$ is then constructed as follows. First, exemplar information is transferred to the scale-specific features by a sequence of cross-attention operations,  
\begin{equation}  \label{eq:cross_attn}
    \mathbf{Q'}_L^{(i)} = \text{CA}(\mathbf{Q'}_L^{(i-1)}, \mathbf{p}_L, \mathbf{p}_L),
\end{equation}
where $\text{CA}(\mathbf{q}, \mathbf{k}, \mathbf{v})$ is the 
standard cross-attention \cite{attention} with query/key/value tensors $(\mathbf{q}, \mathbf{k}, \mathbf{v})$, including a skip connection, followed by a layer normalization, with  
$\mathbf{Q'}_L^{(0)}=\mathbf{C}_\text{L}$ and $i=1:N_{CA}$.

The queries are then refined by a sequence of local transformations, employing deformable attention DA($\cdot$)~\cite{deformable-detr}, which attends to only a small number of nearby queries with a learnable displacement prediction. 
This enables processing low- and high-resolution scales without quadratic computational complexity restrictions of full self-attention, i.e., 
 
\begin{equation}  \label{eq:deform_attn}
   \mathbf{Q''}_L^{(j)} = \text{DA}(\mathbf{Q''}_L^{(j-1)}),
\end{equation}
where $\mathbf{Q''}_{L}^{(0)} = \mathbf{Q'}_L^{(N_{CA})}$ and $j=1:N_{\text{DA}}$. 
The output of the scale-specific query encoder at $L$-th scale is thus $\mathbf{Q}_L = \mathbf{Q''}_L^{(N_{\text{DA}})}$.

\subsection{Cross-Scale Query Aggregation}  \label{sec:scale_aggregation}
 
The scale-specific object queries $\mathbf{Q}_L$ are gradually aggregated into generalized-scale high-resolution dense object queries $\mathbf{Q}$. 
The aggregation is performed by consecutive lightweight upsampling and fusion modules (denoted as LUM), each composed of $2\times$ bilinear upsampling, $3\times3$ convolution, and a GeLU activation. 
Starting from the coarsest resolution, queries are progressively generalized across scales to a higher resolution, i.e.,
 
\begin{equation}  \label{eq:LUM}
   \mathbf{Q} = \mathrm{LUM}_3\Bigl( \mathrm{LUM}_2\bigl( \mathrm{LUM}_1( \mathbf{Q}_1) + \mathbf{Q}_2\bigl) + \mathbf{Q}_3\Bigl).
\end{equation}
Note that separate trainable parameters are used in each LUM$_L$, to enable scale-specific fusion.
Since we consider three scales, the output dense object query map from the cross-scale query aggregation module is 
$\mathbf{Q} \in \mathbb{R}^{H \times W \times d}$, where $H=H_0/2$, $W=W_0/2$.

\subsection{Dense Query Decoder}  \label{sec:query_decoder}
 
The dense query decoder transforms the generalized-scale high-resolution dense object queries $\mathbf{Q}$ into object detections.
In particular, two heads are employed to predict dense objectness score map and per-location bounding boxes.
The objectness score map $\mathbf{y}^{\mathrm{O}} \in \mathbb{R}^{HW \times 1}$ is obtained by applying a linear-layer transformation, followed by a Leaky ReLU activation,
 
\begin{equation} \label{eq:y_pred}
\mathbf{y}^{\mathrm{O}} = \text{LReLU}(\mathbf{W}^O \mathbf{Q}), 
\end{equation}
where $\mathbf{W}^O$ is a learned matrix. 
The bounding boxes $\mathbf{y}^{\mathrm{BB}} \in \mathbb{R}^{HW \times 4}$ are estimated in the {\it tlrb} format~\cite{tian2019fcos} using a three-layer MLP,

\begin{equation} \label{eq:bb_pred}
\mathbf{y}^{\mathrm{BB}} = \sigma(\text{MLP}(\mathbf{Q})), 
\end{equation}
where $\sigma(\cdot)$ denotes the sigmoid function. 
Note that $\mathbf{Q}$ in~(\ref{eq:y_pred}) and~(\ref{eq:bb_pred}) is reshaped into $HW \times d$ for computation purposes.
The final detections $\boldsymbol{B}^{\text{P}}$ are obtained from $\mathbf{y}^{\mathrm{O}} \in \mathbb{R}^{H \times W \times 1}$ and $\mathbf{y}^{\mathrm{BB}} \in \mathbb{R}^{H \times W \times 4}$ as follows. 
Bounding boxes on spatial locations with objectness scores larger than their immediate 8 neighbors (a local maximum), surpassing a threshold set on the training set, are retrieved from $\mathbf{y}^{\mathrm{BB}}$.
During inference, these bounding boxes are fed as prompts to SAM2~\cite{sam2} decoder on the already computed backbone 
features $\mathbf{f}^{I}$ to predict segmentation masks. 
Each bounding box is then refined by min-max fit to the corresponding mask. 
Note that in cases with over 800 objects detected, the average object size is approximately 25 pixels, at which point SAM2 masks become unreliable and are ignored.
Finally, a non-maxima suppression is applied to the bounding boxes to remove duplicate detections, yielding the output bounding boxes $\mathbf{B}^{P}$ and masks $\mathbf{M}^{P}$.

\subsection{Training}
\label{sec:training}
 
The recent detection-oriented counting loss $\mathcal{L'}$ with automatic hard-negative mining on $\mathbf{y}^{\mathrm{O}}$ and $\mathbf{y}^{\mathrm{BB}}$ is applied~\cite{geco}, without the need for initial warm-up period, which we attribute to the new dense query map robustness.
To enhance small object detection, an auxiliary loss computed from the highest feature resolution level is added. 
The dense object queries $\mathbf{Q}_3$ are transformed by a separate LUM module, i.e., $\mathbf{Q}^\text{AUX} = \text{LUM}_\text{AUX}(\mathbf{Q}_3)$, decoded by a separate Query Decoder, and a detection-oriented counting loss $\mathcal{L'_\text{AUX}}$ is computed.
The auxiliary loss is computed only on images with average object size lower than $\theta_\text{size}=25$ pixels. The final loss is thus $\mathcal{L} =  \mathcal{L'} + \alpha \mathcal{L'}_{\text{AUX}}.$

\section{Experiments}
 
\subsection{Implementation details}
 
\name{} uses a pre-trained Hiera backbone~\cite{hiera} from SAM2~\cite{sam2}, which represents a unique selection due to its high-resolution multi-scale output. The backbone encodes the input image into three levels of features, all with channel depth $d=256$.
During training, the number of exemplars is set to $k=3$. 
In the Scale-Specific Query Encoder, \name{} performs $N_{\text{CA}}=3$ iterations of prototype-to-image interactions, and $N_\text{DA} = 2$ query refinement iterations. 
\name{} follows the standard test-time practice, where the input image is scaled to fit the width and height of the average exemplar under 80 pixels. The image is then zero-padded to $W_0 = H_0 = 1024$, which is the \textit{constant} input resolution to \name{}.

The Hiera backbone parameters are frozen during training, while the other parameters of \name{} are trained for 
200 epochs using the AdamW optimizer~\cite{loshchilov2017decoupled} with a batch size of 8, the initial learning rate of $10^{-4}$, and a weight decay of $5\times10^{-5}$. The auxiliary loss weight is set to $\alpha = 0.3$.
Training runs on two A100 GPUs with the standard scale augmentations~\cite{dave, djukic_loca, geco} and images zero-padded to $1024 \times 1024$ pixels.

\subsection{Experimental protocols and benchmarks}
 
\name{} is evaluated on two well-established few-shot counting and detection benchmarks FSCD147~\cite{counting-detr} and FSCD-LVIS~\cite{counting-detr}, and the most recent MCAC~\cite{mcac}, a multiclass synthesized benchmark.
FSCD147~\cite{counting-detr} extends FSC147~\cite{famnet}, a dataset comprising 6135 images across 147 object categories, with bounding box annotations. The dataset is split into 3659 training, 1286 validation, and 1190 test images, with disjoint categories ensuring that objects in the test set are unseen during training. 
FSCD-LVIS~\cite{counting-detr} features 377 object categories. 
We use the unseen split, which contains 3959 training and 2242 test images, ensuring no overlap in object categories between training and testing.
The Multi-class, Class-Agnostic Counting dataset MCAC~\cite{mcac}, contains 1–4 object classes per image.
For every preset category, separate counting runs are conducted, ensuring that models are rigorously tested on their ability to specialize at inference time for the target category.

Counting performance is evaluated by following the standard protocol~\cite{famnet}, 
using Mean Absolute Error (MAE) and Root Mean Squared Error (RMSE). 
Detection accuracy is measured by Average Precision (AP) and AP at IoU=50 (AP50)~\cite{counting-detr}. 
Both GPU usage and inference time are tracked using PyTorch on a single A100, measuring reserved memory, and execution time, excluding data loading. For additional insights, ablations and evaluations, please see the supplementary material.

\begin{table*}[h!]
\centering
\caption{
Performance of few-shot global counting methods (top part) and detection-based methods (bottom part) on FSCD147.
}
\label{tab:fsc147-results}
{\small
\begin{tabular}{lllllllll}
\toprule
                             & \multicolumn{4}{c}{Validation set}                                          & \multicolumn{4}{c}{Test set}                                               \\  \cmidrule(lr){2-5} \cmidrule(lr){6-9}
Method                       & MAE ($\downarrow$) & RMSE($\downarrow$) & AP($\uparrow$) & AP50($\uparrow$) & MAE($\downarrow$) & RMSE($\downarrow$) & AP($\uparrow$) & AP50($\uparrow$) \\ \midrule
FamNet
& 23.75              & 69.07              & -              & -                & 22.08             & 99.54              & -              & -                \\
CFOCNet
& 21.19              & 61.41              & -              & -                & 22.10             & 112.71             & -              & -                \\
BMNet+
& 15.74              & 58.53              & -              & -                & 14.62             & 91.83              & -              & -                \\
VCN
& 19.38              & 60.15              & -              & -                & 18.17             & 95.60              & -              & -                \\
SAFEC
& 15.28              & 47.20              & -              & -                & 14.32             & 85.54              & -              & -                \\
CounTR
& 13.13              & 49.83              & -              & -                & 11.95             & 91.23              & -              & -                \\
LOCA
& 10.24              & 32.56              & -              & -                & 10.79             & 56.97              & -              & -                \\

CACViT
& 10.63              & 37.95              & -              & -                & 9.13 & 48.96              & -              & -                \\ 
CountGD
& 8.12              & 38.97              & -              & -                & 8.35             & 89.80              & -              & -                \\ 
\arrayrulecolor[gray]{0.6}
\midrule
\arrayrulecolor[gray]{0.0}
\rowcolor[gray]{0.975}
C-DETR
& 20.38              & 82.45              & 17.27          & 41.90            & 16.79             & 123.56             & 22.66          & 50.57            \\
\rowcolor[gray]{0.975}
SAM-C
& 31.20              & 100.83             & 20.08          & 39.02            & 27.97             & 131.24             & 27.99         & 49.17            \\
\rowcolor[gray]{0.975}
PSECO
& 15.31      & 68.36      & 32.12\third{}             & 60.02                & 13.05    & 112.86     & 42.98\third{}        & 73.33\third{}           \\
\rowcolor[gray]{0.975}
DAVE
& 9.75\third{}      & 40.30\third{}      & 24.20         & 61.08\third{}           & 10.45\third{}    & 74.51\third{}     & 26.81          & 62.82            \\
\rowcolor[gray]{0.975}
GeCo
& 9.52\second{}      & 43.00\second{}    & 33.51\second{}          & 62.51\second{}           & 7.91\second{}      & 54.28\second{}     & 43.42\second{}         & 75.06\second{}           \\ 
\rowcolor[gray]{0.975}
\name{}                         & \textbf{9.40}\first{}      & \textbf{33.28}\first{}    & \textbf{34.08}\first{}          & \textbf{63.21}\first{}           & \textbf{7.64}\first{}      & \textbf{39.39}\first{}     & \textbf{46.81}\first{}         & \textbf{75.89}\first{}           \\ 

\bottomrule
\end{tabular}}

\end{table*}

\subsection{State-of-the-Art Comparison}
 
\name{} is compared with the currently best state-of-the-art detection-based counters C-DETR~\cite{counting-detr}, SAM-C~\cite{samcount}, PSECO~\cite{pseco}, DAVE~\cite{dave}, GeCo~\cite{geco}, which provide object locations by bounding boxes, and CountGD~\cite{countgd} that provides point localization if test-time normalization or tiling is not activated. For completeness, we include state-of-the-art global counters, which only estimate the total count without localization, i.e., LOCA~\cite{djukic_loca}, CACViT~\cite{cacvit}, CounTR~\cite{Liu_2022_BMVC}, SAFECount~\cite{you2023few}, BMNet+~\cite{Shi_2022_CVPR},  VCN~\cite{Ranjan_2022_CVPR}, CFOCNet~\cite{yang2021class}, MAML~\cite{finn2017model}, FamNet~\cite{famnet} and CFOCNet~\cite{yang2021class}.
Results are summarized in Table~\ref{tab:fsc147-results}\footnote{Detailed qualitative analysis in the supplementary material}.

\name{} outperforms all state-of-the-art detection-based counters on the FSCD147~\cite{counting-detr} dataset, including the most recent GeCo~\cite{geco}, with improvements of 22\% and 27\% in RMSE on the test and validation splits, respectively, and 3\% MAE on the test split (see Table~\ref{tab:fsc147-results}). 
Notably, \name{} is the only detection-based few-shot counter that surpasses all global counters, outperforming CACViT by 16\% MAE and by a remarkable 20\% RMSE on the test set. 
This advancement is driven by its scale-generalized query formulation, effectively addressing the challenges in dense regions with small objects, without resorting to ad-hoc scaling heuristics like the current state-of-the-art.
\name{} achieves superior detection accuracy, outperforming all state-of-the-art methods, e.g., the three-stage PSECO~\cite{pseco}, which leverages both SAM~\cite{sam} and CLIP~\cite{clip}, and two-stage DAVE~\cite{dave}, which employs an additional verification step. 
Furthermore, the current state-of-the-art GeCo~\cite{geco}, which also exploits SAM, is outperformed by 8\% AP on FSCD147 dataset~\cite{counting-detr}. 
By excelling in both total count estimation as well as in localization accuracy, \name{} demonstrates its superiority in low-shot counting.

\begin{table}
    \centering
    \captionof{table}{Few-shot counting and detection on the FSCD-LVIS ''unseen'' split.}
    \label{tab:fscd-lvis-results}
{\small
    \begin{tabular}{llllll}
        \toprule
        & \multicolumn{2}{c}{Count} & \multicolumn{2}{c}{Detection} \\
        \cmidrule(lr){2-3} \cmidrule(lr){4-5}
         Method & MAE($\downarrow$)  & RMSE($\downarrow$)  & AP($\uparrow$)  & AP50($\uparrow$)  \\
        \midrule
        AttRPN-PB
        & 39.16 & 46.09& 3.15 & 7.87 \\
        C-DETR
        & 23.50 & 35.89& 3.85 & 11.28 \\
        DAVE
        &15.47\third{}&25.95\second{} & 4.12\third{} & 14.16\third{} \\
        GeCo
        &15.26\second{}&28.80\third{} &11.47\second{} & 24.49\second{}\\
        \name{} &\textbf{13.01}\first{} &\textbf{23.43}\first{} &\textbf{14.08}\first{} & \textbf{28.84}\first{}\\
        \bottomrule
    \end{tabular}}

\end{table}

\name{} is further evaluated on FSCD-LVIS~\cite{counting-detr}, unseen split, which follows the few-shot counting setup, where categories across splits are disjoint. Results in Table~\ref{tab:fscd-lvis-results} show that \name{} outperforms all state-of-the-art detection counters by 15\% MAE and 19\% RMSE in counting, but also by 23\% AP and 18\% AP50 in detection. These results on the detection-field-adapted FSCD-LVIS support the conclusions drawn from the FSCD147 experiments.\footnote{Evaluation on CARPK in the supplementary material.}

\textbf{Multi-Class Performance.} 
As observed in~\cite{dave,ciampi}, state-of-the-art methods often prioritize high recall, leading to overcounting by indiscriminately detecting all objects in the image, even categories not specified by the exemplars. 
To further evaluate \name{}, we test it on a recently proposed synthetic multi-class dataset MCAC~\cite{mcac}, where 52\% of the images contain multiple object categories, requiring separate count predictions for each category.
\name{} is compared against state-of-the-art global counters from~\cite{ciampi}, including LOCA~\cite{djukic_loca}, CounTR~\cite{Liu_2022_BMVC}, BMNet+\cite{Shi_2022_CVPR}, and FamNet\cite{famnet}.
Results in Table~\ref{tab:mcac} show that \name{} outperforms all few-shot counters, surpassing the benchmark top-performer, LOCA, by 27\% in MAE and 22\% in RMSE on the test set. These results indicate that \name{} successfully differentiates between object categories, a critical ability that is overlooked~\cite{dave,ciampi} in experimental evaluation on FSC147~\cite{famnet}.
\begin{table}[h!]
    \centering
    \captionof{table}{Few-shot counting on MCAC benchmark.}
    \label{tab:mcac}
{\small
    \begin{tabular}{llllll}
        \toprule
        & \multicolumn{2}{c}{Validation} & \multicolumn{2}{c}{Test} \\
        \cmidrule(lr){2-3} \cmidrule(lr){4-5}
         Method & MAE($\downarrow$)  & RMSE($\downarrow$)  & MAE($\downarrow$)  & RMSE($\downarrow$)  \\
        \midrule
        Famnet+
        & 24.76         & 41.12          & 26.40         & 45.52          \\
        Bmnet+
        & 15.83         & 27.07          & 17.29         & 29.83          \\
        GeCo  &   15.31   &   28.42    &  19.91  &33.25       \\
        CounTR
        & 15.07\third{}         & 26.26\third{}          & 16.12\third{}         & 29.28\third{}          \\
        LOCA
        & 10.45\second{}         & 20.81\second{}           & 10.91\second{}  & 22.04\second{}             \\
        \name{}   &  \textbf{9.38}\first{}    &   \textbf{18.80}\first{}    &  \textbf{7.93}\first{}   &  \textbf{17.05}\first{}     \\
        \bottomrule
    \end{tabular}
    }
\end{table}

\subsection{Comparison with scaling-heuristic-free sota}

As already mentioned, the current best-performing few-shot counters rely on ad-hoc resolution upscaling strategies. 
In particular
GeCo~\cite{geco} upscales the input images with small objects to 1536$\times$1536, whereas CountGD~\cite{countgd} performs tiled inference, raising the effective resolution up to 3465$\times$2400. 
These techniques are crucial for achieving high performance, but they significantly increase computation and GPU memory requirements, thereby constraining their applicability.
In contrast, \name{} performs inference on fixed input resolution of 1024$\times$1024, with all images zero-padded if needed to fit this size. 
To emphasize the object-size generalization issues of existing methods, we turned off their ad-hoc scaling heuristics. 

\begin{table}
    \centering
    \captionof{table}{
        Few-shot counters on the FSCD147 
         without upscaling and tiling. Performance drops are marked in red.
    }
    \label{tab:heuristic-free}
\resizebox{0.47\textwidth}{!}{
    \begin{tabular}{lllll}
    \toprule
    Method & MAE ($\downarrow$) & RMSE ($\downarrow$) & AP ($\uparrow$) & AP50 ($\uparrow$) \\
    \midrule
    PSECO
    & 13.05 & 112.86 & 42.98 & 73.33 \\
    LOCA
    & 10.79 & 56.97 & - & - \\
    CountGD
    & 9.44 \textcolor{red}{\small{↓1.0}} & 98.84\textcolor{red}{\small{↓9.0}} & - & - \\
    DAVE
    & 12.21\textcolor{red}{\small{↓1.8}} & 94.93\textcolor{red}{\small{↓20.4}} & 26.44\textcolor{red}{\small{↓0.4}} & 61.04\textcolor{red}{\small{↓1.8}} \\
    GeCo
    & 8.88 \textcolor{red}{\small{↓1.0}} & 75.67\textcolor{red}{\small{↓21.4}} & 43.15\textcolor{red}{\small{↓0.3}} & 73.40\textcolor{red}{\small{↓1.7}} \\
    \name & \textbf{7.64} & \textbf{39.4} & \textbf{46.81} & \textbf{75.89} \\
    \bottomrule
    \end{tabular}}
\end{table}
The results are shown in Table~\ref{tab:heuristic-free}. 
\name{} maintains its performance, since it does not apply the heuristics.
On the other hand, the current state-of-the-art 
performance falls below the classic density-based counter LOCA~\cite{djukic_loca}, uncerscoring their significant reliance on image upscaling. The performance gap between \name{} and GeCo on the FSCD147~\cite{famnet} test set further increases to 13\% and 48\% in MAE and RMSE, respectively. Additionally, \name{} surpasses CountGD by 19\% in MAE and a significant 151\% in RMSE.

The observed RMSE drop of 39\% for GeCo~\cite{geco}, 27\% for DAVE~\cite{dave}, and 10\% for CountGD highlights the poor generalization of state-of-the-art methods across different scales. 
In contrast, GeCo2 handles multiple scales in a principled way without reliance on ad-hoc pre-processing and without computational overhead.

\subsection{Memory and inference speed}
 
To assess the practical applicability of few-shot counters, we measured the average inference time per any image and the average inference time per image for images containing at least 300 objects on the validation and test set from FSCD147~\cite{famnet}, and reported the MAE error for this subset of dense scenes images.
Additionally, we recorded the per-image GPU memory usage when ad-hoc scaling or tiling is applied in the counters. 

The results are summarized in Table~\ref{tab:complexity}.
Related methods require significantly more resources than \name{}, consuming at least 210\% more GPU memory. On average, \name{} infers the total count in 0.4 seconds, whereas the recent GeCo~\cite{geco} and CountGD~\cite{countgd} are 3$\times$ and 6$\times$ slower, respectively. 
Note that the gap between \name{} and other counters increases even further on images with more than 300 objects, when the average width/height of objects is $\sim$50 px. 
In particular, GeCo~\cite{geco} and CountGD~\cite{countgd} become 6$\times$ and 8$\times$ slower than \name{}. 

To further analyze the source of the performance efficiency gain for \name{}, we measure 
the average inference times, excluding the backbones. We observe a 3.8$\times$ speedup for \name{} (0.13 s) compared to GeCo (0.49 s). 
This conclusively validates the efficiency of the new multi-scale query map construction that eliminates the 
need for ad-hoc pre/post processing, while leading to a 
substantial $\sim$~10\% error reduction in challenging dense scenarios.

To isolate the effect of the backbone, we trained GeCo~\cite{geco} with the Hiera backbone from SAM2, denoted GeCo+Hiera (see Table~\ref{tab:ablation}). This variant requires 17GB RAM and averages 0.3s per FSCD147 image--1.5$\times$ slower than \name{}. On dense scenes, inference time increases to 1.9s, nearly 5$\times$ slower, confirming that the efficiency gains of \name{} stem from the new query map construction method, rather than from the backbone.

\begin{table}
    \centering
    \captionof{table}{
        Inference time and GPU memory per image on FSCD147; $\dagger$ for scenes with over 300 objects.
    }
    \label{tab:complexity}
{\small
    \begin{tabular}{lllll}
    \toprule
    Method & mem(Gb) & time(s) & time(s)$\dagger$ & MAE$\dagger$ \\
    \midrule
    DAVE
    & 25.33 & 0.3 \textcolor{red}{\small{↓1.5×}} & 1.4 \textcolor{red}{\small{↓3.5×}} & 113.88 \\
    CountGD
    & 15.47 & 1.3 \textcolor{red}{\small{↓6.5×}} & 3.3 \textcolor{red}{\small{↓8.3×}} & 138.99 \\
    GeCo
    & 23.83 & 0.6 \textcolor{red}{\small{↓3×}} & 2.3 \textcolor{red}{\small{↓5.8×}} & 108.13 \\
    \name{} & 4.98 & 0.2 & 0.4 & 96.92 \\
    \bottomrule
    \end{tabular}}
\end{table}

\subsection{Ablation Study}

\begin{figure}[h!]
    \centering
    \includegraphics[width=0.99\columnwidth]{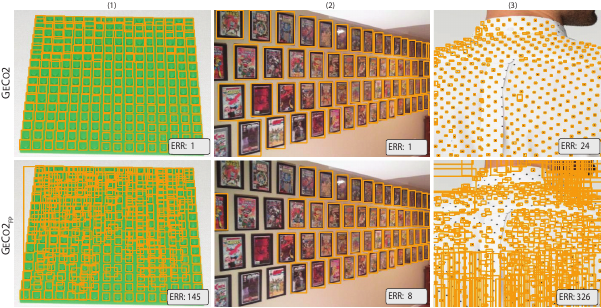}
    \caption{\name{} and \name{}$_{FP}$ with removed multi-scale query generation, using $\mathbf{C}_\text{2}$ and $\mathbf{C}_\text{3}$ for upsampling.}
    \label{fig:ablation}
\end{figure}

Ablation results on FSC147~\cite{famnet} are summarized in Table~\ref{tab:ablation}.
First, to assess the contribution of the Scale-Specific Query Encoder, we ablate the multi-scale query formulation by limiting the prototype-image interaction only to the lowest scale $Q_1$. 
The query generation process can be defined by rewriting (\ref{eq:LUM}) as follows
\begin{equation}  \label{eq:LUM_abl}
   \mathbf{Q} = \mathrm{LUM}_3\Bigl( \mathrm{LUM}_2\bigl( \mathrm{LUM}_1( \mathbf{Q}_1) + \mathbf{C}_2\bigl) + \mathbf{C}_3\Bigl).
\end{equation}
This version (denoted \name{}$_{FP}$) can be viewed as a common pipeline with interaction performed on the lowest scale, followed by a feature pyramid to get more datails from high-resolution features. 
This variant incurs a substantial 11\% increase in MAE and 51\% in RMSE. 
The qualitative comparison in Figure~\ref{fig:ablation} further confirms this degradation, demonstrating the necessity of query generation at multiple scales. The proposed multi-scale query formulation is key to robust scale generalization, enabling accurate detections of small objects in dense regions (rows 1 and 2) without reliance on ad-hoc upscaling strategies, as well as objects with varying sizes within a single image (row 3). 
Furthermore, to demonstrate the impact of individual feature layers, \name{} is modified to use only features from the first, second, or third backbone layers. 
These versions are denoted by \name{}$_\mathbf{Q_1}$ \name{}$_\mathbf{Q_2}$ and \name{}$_\mathbf{Q_3}$, respectively. 
The results clearly show that both versions result in a significant performance drop in both counting and detection, which implies that using only a single level of high-resolution features does not deliver good performance. 
Deep low-resolution features capture more semantic context and perform well in complex scenes, whereas, high-resolution features are critical for detecting small objects and handling densely populated regions. The proposed query map construction leverages both: it individually generates scale-specific queries, which are progressively refined across scales, enabling strong semantic reasoning without sacrificing spatial precision.

\label{sec:ablation}

We also ablate $N_{\text{CA}}$, which when set to $N_{\text{CA}} = 2$ results in a minor drop (3\% MAE, 10\% RMSE). Setting number of deformable attentions $N_{\text{DA}}=1$, for query refinement, results in on-pair performance considering MAE and 30\% RMSE performance drop
This justifies the proposed prototype formulation and dense query transformation steps. 

To asses the importance of auxiliary supervision for small objects, we set the auxiliary loss weight $\alpha = 0$. An increase of 8\% in MAE and 35\% in RMSE, along with reduced detection metrics is observed. This underscores the importance of the additional training signal on the highest resolution of features $Q_3$ for guiding the network to learn accurate small object detection.
Alternatively, larger values, e.g., $\alpha = 0.5$, also deteriorate the performance.

To give insight into the shape prototype encoder that is shared over the scales, we re-trained a variant that uses a separate encoder for each scale (\name{}$_{3\Phi(\cdot)}$). This led to 16\% MAE and 60\% RMSE performance drop, confirming that integrating the same shape prototypes at multiple scales is crucial for providing the same object size cues when generating object queries at all feature levels.
Collectively, ablation experiments validate that the architecture design -- comprising multi-scale dense query formulation of \name{}, deformable attention for query refinement, auxiliary supervision for small objects, and object-size-cue-providing shape prototypes -- is designed robustly, and efficiently.

We also evaluate the impact of SAM2~\cite{sam2} mask refinement. Removing it (\name{}$_{\overline{\text{REF}}}$) does not increase global count errors (MAE/RMSE), but results in a 30\% AP and a modest 4\% AP50 drops. 
This means, that although SAM2 indeed improves individual box quality, the objects are already well localized before refinement. 

Lastly, we ablate the effect of the SAM2 backbone by integrating it with GeCo, to verify whether the performance gains can be attributed the better backbone, rather than the proposed multi-scale dense query construction.
The resulting variant, GeCo+Hiera, demonstrates improved detection performance over the original GeCo, yet \name{} consistently outperforms it in both counting and detection metrics. This confirms that the gains of \name{} are primarily driven by its scale-specific query design rather than the backbone substitution.

\begin{table}[h]
 {\small   
    \caption{Ablation study on the FSCD147 val split.
    }
    \label{tab:ablation}
    \centering

    \begin{tabular}{lllll}
        \toprule
 & \multicolumn{2}{c}{Counting} & \multicolumn{2}{c}{Detection} \\
        \cmidrule(lr){2-3} \cmidrule(lr){4-5}
                Method & MAE($\downarrow$)  & RMSE($\downarrow$)& AP($\uparrow$)  & AP50($\uparrow$)  \\
        \midrule
       \name{} & 9.40      & 33.28    & 34.08          & 63.21 \\
       \midrule
       \name{}$_{FP}$ & 10.48 & 50.49 & 34.02 & 62.43 \\
       \name{}$_\mathbf{Q_1}$ & 12.71 & 60.29  & 34.05 & 62.10 \\
       \name{}$_\mathbf{Q_2}$ & 22.64 & 58.15 & 24.98 & 49.86 \\
       \name{}$_\mathbf{Q_3}$ & 19.97 & 55.33 & 24.75 & 47.97 \\
       \name{}$_{N_{\text{DA}=1}}$ & 9.45& 43.19 & 34.03 & 63.05 \\
       \name{}$_{N_{\text{CA}=2}}$ & 11.06 & 45.68 & 33.58 & 62.15 \\
        \name{}$_{\alpha=0}$  & 10.15 & 44.89& 33.61&61.19 \\
        \name{}$_{\alpha=0.5}$  &9.39 & 36.74&33.47 &63.02 \\
        \name{}$_{\overline{\text{REF}}}$ & 9.13 & 31.76 & 24.15& 60.22 \\
       \name{}$_{3\Phi(\cdot)}$ &10.93 &53.08 & 34.63&63.82 \\

     GeCo+Hiera & 10.99& 47.99& 33.75&62.84 \\
    
        \bottomrule
    \end{tabular}
}
\end{table}

\section{Conclusion}
 
We proposed \name{}, a novel end-to-end low-shot detection-based object counter that explicitly models scale through exemplar-conditioned, scale-specific query encoders and a generalized-scale aggregation strategy, all while maintaining a low memory footprint.
Unlike other counters that rely on computationally expensive upsampling or tiling, \name{} delivers accurate counting even in dense scenes.
Extensive evaluations on multiple benchmarks show that \name{} outperforms state-of-the-art detection-based and global counters, reducing MAE by $\sim$10\% and RMSE by $\sim$20\%. 
Furthermore, \name{} surpasses detection-based counters by $\sim$10\% AP while running 3$\times$ faster with a lower memory footprint. 
By eliminating the ad-hoc heuristics, \name{} provides a principled architecture and sets a new state-of-the-art in few-shot counting.

\name{} relies on backbones that supports high-resolution input and yield high-quality multi-scale features, which may be seen as its limitation. Yet, it naturally aligns it with the SAM2 Hiera~\cite{sam2} encoder, which positions \name{} as an additional head on SAM2 swiss-army knife toolbox, seamlessly extending SAM2 with unprecedented counting capabilities.

{\footnotesize
\noindent\textbf{Acknowledgements.}
This work was supported by the Slovenian Research Agency program P2-0214 and projects L2-3169, J2-60054, as well as the supercomputing network SLING (ARNES, EuroHPC Vega - IZUM).
}

\bibliography{aaai2026}

\end{document}